\documentclass[runningheads]{llncs}
\usepackage[utf8]{inputenc}

\usepackage{amsmath}
\usepackage{amssymb}
\usepackage{tabularx}

\usepackage{graphicx}
\usepackage{algorithmic}
\usepackage{algorithm}
\usepackage{pgfplots}
\pgfplotsset{compat=1.16}
\usepackage{pgffor}

\usepackage{listings}
\usepackage[mathscr]{euscript}
\let\euscr\mathscr \let\mathscr\relax
\usepackage[scr]{rsfso}

\usepackage[most]{tcolorbox}
\newtcbinputlisting[auto counter,list inside=lol,list type={lstlisting}]{\mylisting}[3][]{%
  listing file={#3},
  title=Listing,
  colback=white,
  colframe=gray!75!black,
  fonttitle=\bfseries,
  listing only,
  breakable,
  title={Sampling Mathematical Expressions},
  listing options={basicstyle=\ttfamily},
  #1
}

\begin{document}

\title{Monte Carlo Search Algorithms Discovering \\
Monte Carlo Tree Search Exploration Terms}
\titlerunning{Monte Carlo Search Algorithms Discovering Exploration Terms}

\author{Tristan Cazenave}

\institute{LAMSADE, Université Paris Dauphine - PSL, CNRS, Paris, France}

\maketitle

\begin{abstract}
Monte Carlo Tree Search and Monte Carlo Search have good results for many combinatorial problems. In this paper we propose to use Monte Carlo Search to design mathematical expressions that are used as exploration terms for Monte Carlo Tree Search algorithms. The optimized Monte Carlo Tree Search algorithms are PUCT and SHUSS. We automatically design the PUCT and the SHUSS root exploration terms. For small search budgets of 32 evaluations the discovered root exploration terms make both algorithms competitive with usual PUCT.
\end{abstract}

\section{Introduction}

Monte Carlo Tree Search \cite{Coulom2006,Kocsis2006} is a well known family of algorithms that were designed for the game of Go and then applied to many different combinatorial problems \cite{BrownePWLCRTPSC2012,swiechowski2023monte}.

Our goal in this paper is to use Monte Carlo Search to improve Monte Carlo Tree Search. This is part of a longstanding goal of using Artificial Intelligence to improve Artificial Intelligence \cite{pitrat2008step}.

The usual way to design an exploration term is to make a theoretical analysis. We take another empirical approach. We randomly generate many exploration terms and keep the ones that work well in practice. This is a simpler approach, yet it can find exploration terms that work well in practice and that surpass the ones found with a theoretical analysis. 

Another approach to the automatic improvement of Monte Carlo Tree Search exploration terms is to use Genetic Programming.  It could evolve Monte Carlo Tree Search algorithms, improving on UCT and RAVE for the game of Go \cite{cazenave2007evolving}. However this approach relies on making the exploration terms play against each other which is time consuming. It is also more complicated than the method we propose in this paper. 

Monte Carlo Tree Search combined with Deep Reinforcement Learning has been used to improve algorithms. AlphaTensor discovered new fast matrix multiplications algorithms playing the tensor game \cite{fawzi2022discovering}. AlphaTensor as well as other Monte Carlo Tree Search algorithms have also been used for quantum circuit optimization  \cite{hummel2022monte,wang2023automated,rosenhahn2023monte,ruiz2024quantum}. New fast sorting algorithms were discovered thanks to Monte Carlo Tree Search with the AlphaDev system \cite{mankowitz2023faster}. 

Monte Carlo Search has been used for discovering mathematical expressions that maximize a given score function \cite{cazenave2010nested,cazenave2013monte}. This was applied to different domains including physics \cite{sun2022symbolic}, finance \cite{cazenave2015forecasting}, and the automated design of functions \cite{illetskova2019nested}.

Refinements of the Monte Carlo Search approach to mathematical expressions discovery include incorporating actor-critic in Monte Carlo Tree Search for symbolic regression \cite{lu2021incorporating}, using a grammar of Monte Carlo Search algorithms \cite{maes2013monte}, controlling the size \cite{moudvrik2017algorithm}, and using GPT as a prior \cite{li2024discovering}.

The automated discovery of optimization algorithms with symbolic program search recently enabled to discover a simple and effective optimization algorithm, Lion (evoLved sIgn mOmeNtum). Lion is more memory-efficient than Adam as it only keeps track of the momentum \cite{chen2024symbolic}.

Our work is in line with these uses of Artificial Intelligence to discover new Artificial Intelligence algorithms. Our goal is to discover new root exploration terms for two Monte Carlo Tree Search algorithms: PUCT \cite{Silver2016MasteringTG} and SHUSS \cite{fabiano2021sequential}. We use Monte Carlo Search to discover these Monte Carlo Tree Search root exploration terms.

Our contributions are:

\begin{itemize}
\item An efficient method to empirically design exploration terms.
\item The AMAF prior for non uniform playouts in Monte Carlo Search applied to the discovery of mathematical expressions.
\item The design of a curriculum learning dataset for discovering exploration terms.
\item A better way of selecting moves for SHUSS according to their priors given by the policy network.
\end{itemize}

The second section presents various Monte Carlo Tree Search algorithms. The third section explains how we generate mathematical expressions for the exploration terms. The fourth section details experimental results.

\section{Monte Carlo Tree Search}

In this section we present various Monte Carlo Tree Search algorithms, starting with PUCT the most popular one which is used in Alpha Zero and that is standard in computer games. We then define the AMAF prior that can be used to play non uniform playouts biased toward the actions that give better playouts scores. It uses statistics on the playouts that contain an action to calculate its probability of being played as explained in the following subsection on sampling. We then present the Sequential Halving algorithm as well as the related Sequential Halving Using Scores (SHUSS) algorithm. We end this section explaining how generated exploration terms can be used for PUCT and SHUSS.

\subsection{PUCT}

\begin{figure}[ht]
    \begin{center}
    \includegraphics [width=7.0cm]  {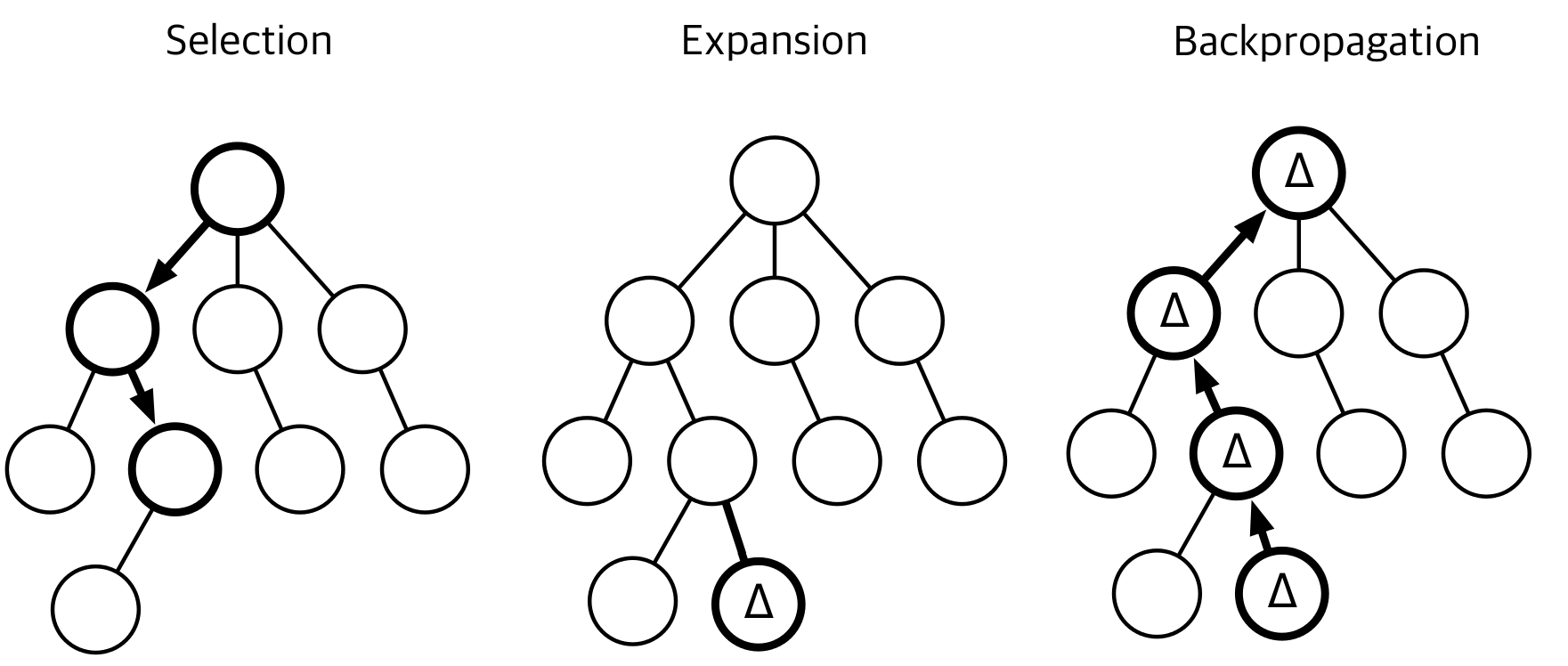}
    \end{center}
    \caption{The three steps of MCTS. The first step is the tree descent using the exploration term to choose among the children. The second step is adding a new leaf associated to an evaluation of the state by the value network. The third step is updating the statistics in the tree with the evaluation.}
    \label{mcts}
    \vspace{2em}
\end{figure}

Monte Carlo Tree Search was designed for computer Go and made a revolution in computer Go \cite{Coulom2006,Kocsis2006} and then in computer game playing \cite{Finnsson2008,mehat2008monte}. The current standard algorithm for Monte Carlo Tree Search is PUCT. This is the search algorithm used in AlphaGo \cite{Silver2016MasteringTG}, AlphaGo Zero \cite{silver2017mastering}, AlphaZero \cite{silver2018general} and MuZero \cite{schrittwieser2020mastering}.

Figure \ref{mcts} gives the three steps of MCTS. The principle of the algorithm is to memorize the already explored states as well as the associated statistics for the possible actions in these states. When the algorithm encounters a state he has already explored, it uses an exploration term to choose the next action to play. The average of the previous evaluations associated to an action $a$ in state $s$ is $Q(s,a)$. The number of descents that have passed by state $s$ is $N(s)$ and the number of descent in $s$ that have played action $a$ is $N(s,a)$.

A neural network is used at the leaves of the tree to evaluate the leaf and calculate probabilities for the possible actions. The probability that action $a$ is the best in state $s$ is $P(s,a)$.

The exploration term used in PUCT is:

$$Q(s,a) + c_p \times P(s,a) \times \frac{\sqrt{N(s)}}{1 + N(s,a)}$$

The constant $c_p$ is an hyper parameter that has to be tuned for each problem.

In the following we use $pr$ for $P(s,a)$ the probability for action $a$ in state $s$ given by the neural network policy head.

\subsection{The AMAF prior}

The All Moves As First (AMAF) heuristic comes from computer Go \cite{Bruegmann1993MC}. It calculates statistics on moves independently of when the moves were played in a playout. It is used in General Game Playing \cite{Pitrat68} in the MAST algorithm \cite{Finnsson2010}. It was also used in computer Go in the RAVE \cite{Gelly2011AI} algorithm. The principle of RAVE is to bias the tree policy with the AMAF statistics of the node. RAVE is much better than UCT for the game of Go. RAVE was later generalized and applied to many games with the GRAVE \cite{cazenave2015generalized} algorithm. GRAVE has much better results than RAVE for many games. It has good results in General Game Playing. It is the standard Monte Carlo Tree Search algorithm used in the Ludii system \cite{browne2020practical}.

We now define a more elaborate version of AMAF. If $\euscr{P}$ is the set of the playouts and $s(p)$ is the score of playout $p \in \euscr{P}$, we define:

$$\mu =  \frac {\sum_{p  \in \euscr{P}}s(p)}{|\euscr{P}|}$$

$$\euscr{P}_a =  \{p  \in \euscr{P}~|~a \in p\}$$

$$\mu_a =  \frac {\sum_{p  \in \euscr{P}_a}s(p)}{|\euscr{P}_a|} - \mu$$

$$maxi = \max_a (\max (\mu_a, -\mu_a))$$

$$z = \sum_a e^{\frac {\mu_a}{maxi}}$$

$$AMAF(a) = \frac {e^{\frac {\mu_a}{maxi}}}{z}$$

\subsection{Sampling}

The basic algorithm in Monte Carlo Search is sampling. It performs playouts by randomly choosing actions until it reaches a terminal state.

It usually improves the results of the playout to adopt a non uniform strategy for sampling. A policy can attribute different probabilities to the possible actions in a state. The sampling algorithm can then choose the next action to play according to these probabilities.

It is also possible to use a temperature $\tau$ to make the policy more or less exploratory. In the case of AMAF, if $\euscr{A}$ is the set of the possible actions in state $s$, sampling with a temperature $\tau$ consists in choosing the next action $a$ with probability $p_a$:

$$p_a = \frac {e^{\tau \times log(AMAF (a))}}{z}$$

$$z = \sum_{a \in \euscr{A}} e^{\tau \times log(AMAF (a))}$$

\subsection{Sequential Halving}

Sequential Halving \cite{SequentialHalving2013} is an algorithm that minimizes the simple regret.
It has successfully been used as an alternative to UCB in Monte Carlo Tree Search, in particular as a replacement in the root node with UCB used in the rest of the tree \cite{pepels2014minimizing}, or even in the whole tree with SHOT \cite{cazenave2015sequential}. It was applied to games as well as to partially observable games \cite{pepels2015sequential}. Sequential Halving was also used as a root policy with Gumbel MuZero \cite{danihelka2021policy}. The outputs of the Sequential Halving at the root were used for reinforcement learning in the MuZero algorithm. Gumbel MuZero was successfully applied to Go, Chess and Atari games.

Algorithm \ref{SH} gives the Sequential Halving algorithm used in SHUSS \cite{fabiano2021sequential}. This is the one we use in this paper with $\lambda = \frac{1}{2}$. The principle of the algorithm is to allocate the same number of playouts to all the actions in the set of actions $S_r$. It then selects half of the actions in $S_r$ that have the best empirical average. This best half constitutes $S_{r+1}$. The algorithm continues to allocate playouts to remaining actions and to select the best half until there is only one action remaining in $S_R$.

\begin{algorithm}
\caption{Sequential Halving}
\begin{algorithmic}
\label{SH}
\STATE {\bfseries \textbf{Parameter}:} cutting ratio $\lambda$
\STATE {\bfseries \textbf{Input}:} total budget $T$, set of arms $S$
\STATE $S_0 \leftarrow S$, $T_0 \leftarrow T$
\STATE $R \leftarrow$ number of rounds before $|S_R|=1$
\FOR{$r=0$ {\bfseries to} $R-1$}
\STATE $t_r \leftarrow \lfloor \frac{T_r}{|S_r| \cdot (R-r)} \rfloor$
\STATE $T_{r+1} \leftarrow T_r - t_r|S_r| $
\STATE sample $t_r$ times each arm in $S_r$ 
\STATE $S_{r+1} \leftarrow S_r$ deprived of the fraction $1-\lambda$ of the worst arms
\ENDFOR
\STATE {\bfseries \textbf{Output}:} arm in $S_R$
\end{algorithmic}
\end{algorithm}

\subsection{Sequential Halving Using Scores}

SHUSS \cite{fabiano2021sequential} is an improvement of Sequential Halving that uses a prior to improve the move selection. The prior can be used either to eliminate moves or to bias the selection of the actions. 

When the prior is standard AMAF it selects the moves to keep using:

$$\tilde{Q}_a = Q_a + C \times \frac{StandardAMAF(a)}{N(root,a)}$$

$$StandardAMAF(a) =  \frac {\sum_{p  \in \euscr{P}_a}s(p)}{|\euscr{P}_a|}$$

In case of a prior given by a neural network, it uses the classic Sequential Halving algorithm restricted to a fixed number of moves that have the best priors.

\subsection{Using an exploration term for the selection of moves in SHUSS}

In the same spirit as SHUSS it is possible to use various exploration terms for choosing the moves to keep at the end of a round of Sequential Halving. Algorithm \ref{Selection} gives the move selection process with an exploration term. The principle is to take the best half of the moves that maximize the expression. If the expression is the usual empirical average then the algorithm is the usual Sequential Halving algorithm.

\begin{algorithm}
\caption{Selection of the moves to keep for the next round}
\begin{algorithmic}
\label{Selection}
\STATE {\bfseries \textbf{Parameter}:} cutting ratio $\lambda$
\STATE {\bfseries \textbf{Input}:} set of moves $S_r$
\STATE $S_{r+1} \leftarrow \emptyset$
\FOR{$i=0$ {\bfseries to} $\lambda \times |S_r|$}
\STATE $bestScore \leftarrow -\infty$
\FOR{$j=0$ {\bfseries to} $|S_r|$}
\IF{$S_r [j] \not \in S_{r+1}$}
\IF{$expression (S_r [j]) > bestScore$}
\STATE $bestMove \leftarrow S_r [j]$
\STATE $bestScore \leftarrow expression (S_r [j])$
\ENDIF
\ENDIF
\ENDFOR
\STATE $S_{r+1} \leftarrow S_{r+1} \cup \{bestMove\}$
\ENDFOR
\RETURN $S_{r+1}$
\end{algorithmic}
\end{algorithm}

\subsection{Using an exploration term for the selection of moves in PUCT}

We also generate exploration terms for PUCT. The generated exploration term is added to the usual PUCT exploration term. Thus the overall exploration term becomes:

$$Q(s,a) + c_e \times P(s,a) \times \frac{\sqrt{N(s)}}{1 + N(s,a)} + expression (a)$$

This exploration term is only used at the root of the PUCT search tree.

\section{Generating Mathematical Expressions}

In this section we detail the algorithms we use to discover mathematical expressions. We first define the expression discovery game, and then explain how to sample expressions for this game.

\subsection{The expression discovery game}

Expression trees are represented as stacks in reverse polish notation. For example the generated expression [+, pr, *, *, 2, sc, sc] corresponds to the exploration term $pr + 2 \times sc \times sc$ where $sc$ is the sum of the scores of the playouts starting with the move to evaluate and $pr$ the prior for that move. The evaluation of an expression in reverse polish notation is algorithmically simple as it uses a straightforward depth first search. 

In order to limit the size of the generated expressions, we maintain the number of open leaves of an incomplete expression. This is the total number of children of the atoms of the expression that are not yet associated to an atom. In the root node the number of open leaves of the empty expression is 1. If for example a '+' atom is assigned to the root, then there is one open leaf less due to the assignment and two open leaves more due to the '+' having two children not yet assigned. In order to generate expressions that are smaller than the maximum length, the legal moves function does not return atoms that make the number of already assigned atoms plus the number of open leaves plus the number of children of the atom greater than the maximum length.

The atoms we used to generate expressions are:

\begin{itemize}
    \item 1, 2, 3 and 100 numbers.
    \item sc: the sum of the scores of the playouts starting with the move.
    \item pr: the prior for the move given by the policy head.
    \item nbp: the number of playouts starting with the move.
    \item nb: the total number of playouts.
    \item +, -, *, /, log, exp, =, max and min operators.
\end{itemize}

\subsection{Sampling}

The default sampling procedure is uniform sampling. It is possible to replace it with non uniform sampling using the AMAF prior. 

In our code for sampling mathematical expressions, the possible atoms are defined in a list and the number of children of each atom is defined in the corresponding children list. A state is a possibly incomplete mathematical expression in reverse polish notation. It is associated to a number of open leaves which is the minimum number of atoms that have to be added to the expression in order to have a complete expression. 

The usual functions to define a problem for Monte Carlo Search are defined as follows for the mathematical expression discovery game:

\begin{itemize}
    \item The legal moves function takes as parameters an incomplete expression and the number of associated open leaves. It returns the list of atoms that can be added to the incomplete expression. It verifies that adding an atom does not exceed the maximal number of atoms for the final complete expression. 
    \item The play function just adds the selected atom to the expression and also returns the updated number of open leaves. 
    \item The terminal function returns True when the expression is complete. 
    \item The playout function is the usual uniformly random playout function that randomly adds authorized atoms to the expression until the expression is complete.
\end{itemize}

The Python code we use for uniform sampling is:\\
\mylisting{}{\jobname.txt}

\section{Experimental Results}

In this section we experiment the discovery of Monte Carlo Tree Search root exploration terms in the game of Go. We present the computer Go dataset that was used to train a transformer network for the game of Go. The transformer network is used to generate the SHUSS dataset that is in turn used to evaluate generated root exploration terms. We compare uniform sampling to AMAF sampling for the discovery of root exploration terms. We apply the framework to the discovery of root exploration terms for both PUCT and SHUSS. We then test the discovered exploration terms in a Go program, making the new algorithms play against standard PUCT with an optimized exploration constant.

\subsection{The computer Go dataset}

The computer Go dataset is composed of games played by Katago \cite{wu2019accelerating} against itself in 2022. There are 1,000,000 different games in total in the training set. The input data is composed of 31 19x19 planes (color to play, ladders, current state on two planes, two previous states on four planes). The output targets are the policy (a vector of size 361 with 1.0 for the move played, 0.0 for the other moves), and the value (close to 1.0 if White wins, close to 0.0 if Black wins). The test set is composed of 50,000 states taken randomly from 50,000 games that are not used in the training set.

\subsection{The neural network}

\begin{figure}[ht]
    \begin{center}
    \includegraphics [width=7.0cm]  {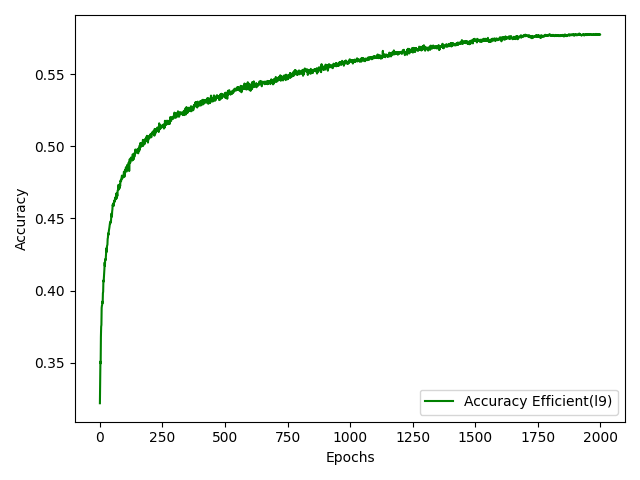}
    \end{center}
    \caption{Evolution of the Accuracy.}
    \label{accuracy}
    \vspace{2em}
\end{figure}

\begin{figure}[ht]
    \begin{center}
    \includegraphics [width=7.0cm]  {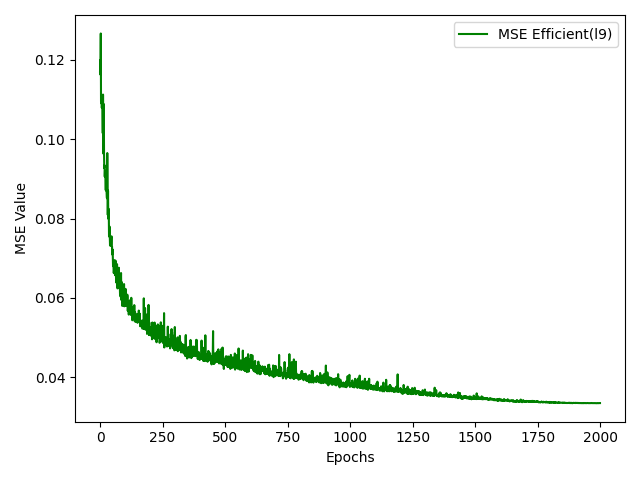}
    \end{center}
    \caption{Evolution of the MSE.}
    \label{mse}
    \vspace{2em}
\end{figure}

\begin{figure}[ht]
    \begin{center}
    \includegraphics [width=7.0cm]  {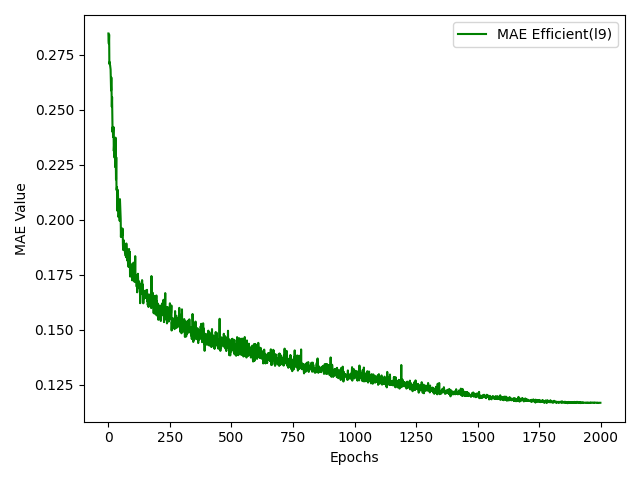}
    \end{center}
    \caption{Evolution of the MAE.}
    \label{mae}
    \vspace{2em}
\end{figure}

We trained a computer Go vision transformer network \cite{sagri2024vision} using 2,000 epochs with 100,000 states per epoch. The loss for the policy head is a categorical cross entropy and the loss for the value head is a binary cross entropy. The optimizer is Adam and the learning rate decreases according to a cosine annealing \cite{cazenave2021cosine}. The network reached an accuracy of 57.75\% on Katago moves as can be seen in Figure \ref{accuracy}, a Mean Squared Error (MSE) of 0.0334 as can be seen in Figure \ref{mse} and a Mean Absolute Error (MAE) of 0.117 as can be seen in Figure \ref{mae}.

\subsection{The SHUSS dataset}

The time required to precisely evaluate a generated expression in a real Go playing program can be huge. For example making a Sequential Halving algorithm with a generated expression play against a standard PUCT for 500 games with 1024 playouts per move takes days.

In order to have a fast evaluation of a given exploration term we built a dataset of states associated to their cached search. The policy learned by the neural network is of high quality. Out of the 2,000 states taken from the test set only 77 of them have a prior less than 0.01. In the remaining of the experiments, we only use moves that have a prior greater or equal to 0.01 for Sequential Halving. For each of the moves that have a prior greater or equal to 0.01, we call PUCT starting with the move a fixed number of times (e.g. 32) and store the sequence of evaluations returned by the successive calls to PUCT after the fixed first move. Therefore for all moves we have an associated sequence of 32 evaluations that can be used to simulate the calls made by Sequential Halving while not using the inference by the neural network. This enables a very fast evaluation of a given Sequential Halving exploration term as the neural network is not used anymore to calculate the accuracy of the exploration term. The accuracy is defined as the number of Katago moves found by the Sequential Halving algorithm using the exploration term on the 2,000 cached states.

We also built a second dataset that uses the same states except that the moves used to calculate the accuracy are the moves found by Sequential Halving with 128 evaluations.

In order to speed up the evaluation of the exploration terms we stop evaluating the accuracy of an exploration term if it scores less than a threshold of 80 after 200 states. We also use memoization of the scores of the exploration terms in a dictionary as well as the memoization of the sums of scores for a given number of playouts and a given first move.

In the experiments we run 100 processes generating exploration terms in parallel for 512 seconds. It results in 354,400 exploration terms being evaluated.

\subsection{AMAF sampling}

\begin{figure}[!b]
    \centering
    \includegraphics [width=7.0cm]  {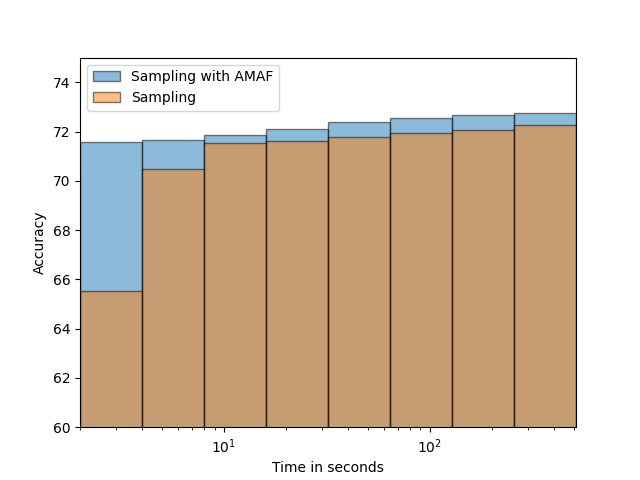}
    \caption{Evolution of the best expression accuracy with the logarithm of the sampling search time with doubling search times. Each measure is the average of 100 runs of the sampling algorithms. Using the AMAF prior improves the results. It finds the same accuracy more than 8 times faster than the uniform sampling algorithm. The temperature of the AMAF sampling is set to 5. The dataset used is the Sequential Halving moves with 128 evaluations dataset and the exploration terms are scored using 32 evaluations on each state out of the 2,000 states.}
    \label{histo}
    \vspace{2em}
\end{figure}

The evolution of the best accuracy on the Sequential Halving moves with 128 evaluations is given in Figure \ref{histo} both for uniform sampling and for AMAF sampling. AMAF sampling is much better.

\begin{table*}[!t]
    \begin{center}
        \begin{tabular}{l|llllllllllll}
            $c_p / c_e$ & 0.05 & 0.10 & 0.15 & 0.20 & 0.25 & 0.30 & 0.35 & 0.40& 0.45 \\
            \hline \\
0.05 & 51.75\% & & 54.00\% & & & & \\
0.10 & & 53.25\% & 53.50\% & & & & \\
0.15 & & & \bf 51.75\% & & & & \\
0.20 & 49.00\% & 47.75\% & 55.75\% & 50.00\% & 48.50\% & 46.00\% & 46.50\% & 43.25\% & 40.25\%\\
0.25 & & & 54.75\% & & 51.50\% & & \\
0.30 & & & 58.00\% & & & 52.00\% & \\
0.35 & & & 60.25\% & & & & 50.75\% \\
0.40 & & & 59.00\% & & & & & 51.00\% &\\
0.45 & & & 63.25\% & & & & & & 50.75\%\\
        \end{tabular}
    \end{center}
    \caption{Winrates against PUCT with 32 evaluations for different exploration constants. The winrates are the winrates for the $\frac{1}{log (sc + nb)}$ discovered exploration term. Each line is the result of 400 games. The starting state for each of the games is a balanced state with the first 20 moves played by Katago. The upper part of the table is used to determine the best $c_p$ value. The lower part is to find the best $c_e$ value against this fixed $c_p$ value. The discovered exploration term has a winrate of 55.75\% against PUCT. The discovered exploration term uses the constant $c_e = 0.15$, and PUCT the constant $c_p = 0.20$. Note that in the upper part of the table $c_p = 0.15$ and $c_e = 0.15$ has a score of 51.00\%. Both of the algorithms can adapt to the other. The discovered exploration term is slightly better.}
    \label{ce}
    \vspace{2em}
\end{table*}

\begin{table*}[!h]
    \begin{center}
        \begin{tabular}{l|llllllllllll}
            $c_p / c_s$ & 0.05 & 0.10 & 0.15 & 0.20 & 0.25 & 0.30 & 0.35 & 0.40& 0.45
            \\ \hline \\
0.05 & 47.50\% & & & & & & \\
0.10 & 43.25\% & 42.75\% & & & & & \\
0.15 & 44.75\% & & 45.75\% & & & & \\
0.20 & 44.75\% & 44.00\% & 41.50\% & 40.25\% & 36.50\% & 39.50\% & 39.50\% & 40.75\% & 41.75\% \\
0.25 & 47.50\% & & & & 40.75\% & & \\
0.30 & \bf 42.50\% & & & & & 41.50\% & \\
0.35 & 47.25\% & & & & & & 45.00\% \\
0.40 & 49.00\% & & & & & & & 50.50\% &\\
0.45 & 54.50\% & & & & & & & & 54.50\%\\
        \end{tabular}
    \end{center}
    \caption{Winrates of standard SHUSS (the exploration term is $sc$) with 32 evaluations and the 5 best prior moves, against PUCT with 32 evaluations for different exploration constants. }
    \label{shuss}
    \vspace{2em}
\end{table*}

\begin{table*}[!h]
    \begin{center}
        \begin{tabular}{l|llllllllllll}
            $c_p / c_s$ & 0.05 & 0.10 & 0.15 & 0.20 & 0.25 & 0.30 & 0.35 & 0.40& 0.45
            \\ \hline \\
0.05 & 53.25\% & & & 57.25\% & & & \\
0.10 & & 52.50\% & & 54.00\% & & & \\
0.15 & 46.25\% & 49.25\% & 47.25\% & 51.50\% & 49.25\% & 44.25\% & 46.25\% & 48.00\% & 46.00\% \\
0.20 & & & & \bf 51.00\% & & & \\
0.25 & & & & 52.75\% & 56.75\% & & \\
0.30 & & & & 56.75\% & & 51.25\% & \\
0.35 & & & & 58.50\% & & & 54.25\% \\
0.40 & & & & 61.00\% & & & & 54.50\% &\\
0.45 & & & & 65.50\% & & & & & 54.50\%\\
        \end{tabular}
    \end{center}
    \caption{Winrates of SHUSS with the $pr + 2 \times sc \times sc$ discovered exploration term and with 32 evaluations and the 5 best prior moves, against PUCT with 32 evaluations for different exploration constants. Using $c_s = 0.20$ always wins against PUCT even if it is only slightly better against PUCT with $c_p = 0.20$. SHUSS with the discovered expression is much better than standard SHUSS.}
    \label{shussPrior}
    \vspace{2em}
\end{table*}

\begin{table*}[!h]
    \begin{center}
        \begin{tabular}{l|llllllllllll}
            $c_p / c_s$ & 0.05 & 0.10 & 0.15 & 0.20 & 0.25 & 0.30 & 0.35 & 0.40& 0.45
            \\ \hline \\
0.05 & 22.25\% & & & & & 24.00\% & \\
0.10 & 19.00\% & 15.50\% & 15.00\% & 16.00\% & 17.25\% & \bf 19.00\% & 17.75\% & 18.25\% & 16.25\% \\
0.15 & & & 18.50\% & & & 19.00\% & \\
0.20 & & & & 22.75\% & & 22.50\% & \\
0.25 & & & & & 25.00\% & 26.75\% & \\
0.30 & & & & & & 29.00\% & \\
0.35 & & & & & & 26.25\% & 27.25\% \\
0.40 & & & & & & 34.75\% & & 31.00\% &\\
0.45 & & & & & & 35.00\% & & & 36.25\%\\
        \end{tabular}
    \end{center}
    \caption{Winrates of SHUSS with the $3 \times pr + max (3, sc)$ discovered exploration term and with 32 evaluations and the 5 best prior moves, against PUCT with 32 evaluations for different exploration constants. The exploration term learned on the Katago moves is also much worse than the exploration term learned on the Sequential Halving moves even though the Katago moves are much better.}
    \label{shussKatago}
    \vspace{2em}
\end{table*}

\begin{table*}[!h]
    \begin{center}
        \begin{tabular}{l|llllllllllll}
            $c_p / c_s$ & 0.05 & 0.10 & 0.15 & 0.20 & 0.25 & 0.30 & 0.35 & 0.40& 0.45
            \\ \hline \\
0.05 & 55.50\% & & 54.25\% & & & & & & \\
0.10 & 47.00\% & 48.75\% & \bf 48.75\% & 46.00\% & 47.25\% & 42.75\% & 41.25\% & 44.25\% & 46.25\%\\
0.15 & & & 49.25\% & & & & & & \\
0.20 & & & 50.00\% & 49.50\% & & & & & \\
0.25 & & & 50.75\% & & 52.50\% & & & & \\
0.30 & & & 52.00\% & & & 53.75\% & & & \\
0.35 & & & 56.25\% & & & & 52.75\% & & \\
0.40 & & & 61.75\% & & & & & 55.75\% & \\
0.45 & & & 61.00\% & & & & & & 55.50\% \\
        \end{tabular}
    \end{center}
    \caption{Winrates of SHUSS with the $pr + 2 \times sc \times sc$ discovered exploration term and with 64 evaluations and the 6 best prior moves, against PUCT with 64 evaluations for different exploration constants. With a different number of evaluations and a different number of best prior moves the discovered exploration term is still competitive with standard PUCT.}
    \label{shuss64}
    \vspace{2em}
\end{table*}

\subsection{Discovering a PUCT exploration term}

Table \ref{PUCT} gives the accuracy of of usual PUCT that uses the $0$ exploration term. It also gives the accuracy of the discovered exploration: $\frac{1}{log (sc + nb)}$. It has a slightly better accuracy than usual PUCT. 

\begin{table}[htbp]
    \begin{center}
        \begin{tabular}{lr}
            \multicolumn{1}{l}{\bf Exploration term}  &\multicolumn{1}{l}{\bf Accuracy}
            \\ \hline \\
    $0$                                  & 72.40\% \\
    $\frac{1}{log (sc + nb)}$~~~~~       & 73.35\% \\
        \end{tabular}
    \end{center}
    \caption{Accuracy on the PUCT moves of PUCT with 32 evaluations using the depicted exploration terms. The sum of the scores of a move is $s$ and the total number of playouts is $nb$.}
    \label{PUCT}
    \vspace{2em}
\end{table}

\subsection{Discovering a SHUSS exploration term}

Table \ref{Katago} gives the accuracy of different exploration terms for the Katago moves. Interestingly the prior $pr$ is much better than the standard Sequential Halving algorithm $sc$ on this dataset. The sampling algorithm finds the $3 \times pr + max (3, sc)$ exploration term that is better than both on this dataset.

Table \ref{SH} gives the accuracy of different exploration terms for the Sequential Halving moves. This time the prior $pr$ is much worse than the standard Sequential Halving algorithm $sc$ on this dataset. The sampling algorithm finds the $sc \times (pr + sc \times sc)$ exploration term that is better than both on this dataset.

\begin{table}[htbp]
    \begin{center}
        \begin{tabular}{lr}
            \multicolumn{1}{l}{\bf Exploration term}  &\multicolumn{1}{l}{\bf Accuracy}
            \\ \hline \\
    $sc$                                  & 44.00\% \\
    $pr$                                  & 56.15\% \\
    $3 \times pr + max (3, sc)$~~~~~      & 57.35\% \\
        \end{tabular}
    \end{center}
    \caption{Accuracy of SHUSS with 32 evaluations on the Katago moves. The score exploration term gives much worse results than the prior exploration term even though the score exploration term is much better for SHUSS. The bias is that the prior is trained on Katago moves. The discovered exploration term reflects this as it gives a lot of importance to the prior.}
    \label{Katago}
    \vspace{2em}
\end{table}

\begin{table}[htbp]
    \begin{center}
        \begin{tabular}{lr}
            \multicolumn{1}{l}{\bf Exploration term}  &\multicolumn{1}{l}{\bf Accuracy}
            \\ \hline \\
    $pr$                                  & 44.85\% \\
    $sc$                                  & 71.45\% \\
    $pr + 2 \times sc \times sc$~~~~~     & 72.85\% \\
        \end{tabular}
    \end{center}
    \caption{Accuracy of SHUSS with 32 evaluations on the SHUSS dataset. The SHUSS label moves are found using Sequential Halving with 128 evaluations. The accuracy is calculated on the moves found by SHUSS with 32 evaluations and the depicted exploration term for halving. The $sc$ exploration term corresponds to standard SHUSS. The $pr + 2 \times sc \times sc$ has a slightly better accuracy on the SHUSS dataset.}
    \label{SH}
    \vspace{2em}
\end{table}

\subsection{Testing the PUCT exploration term in a  Go program}

The method to evaluate the discovered exploration term is to make the program with the discovered exploration term play against PUCT for different values of the exploration constant. The exploration constant for PUCT is noted $c_p$ and the exploration constant for the discovered exploration term is $c_e$. We first make the algorithm play against each other with the same values for $c_p$ and $c_e$ and we retain the best performing constant $c_p$ for PUCT. We then make PUCT with this $c_p$ constant play against the discovered exploration term for various $c_e$ values. It enables to find a good value for $c_e$ against a reasonable value for $c_p$. We finally make PUCT with different values for $c_p$ play against the selected $c_e$ value. The results of this experiment are given in Table \ref{ce}. Each value in the Table requires more than two hours be calculated by playing 400 games, running 50 games in parallel on an EPYC computer. This is why we only do not calculate all the values in the Table. The 400 starting states where taken from the Katago dataset test set games by playing 20 moves by Katago from the beginning of the games. The resulting states are balanced according to Katago. The $sc$ exploration term is the usual exploration term for Sequential Halving. 

\subsection{Testing the SHUSS exploration term in a  Go program}

We evaluate the different exploration terms by using them in a computer Go program. The SHUSS algorithm using an exploration term plays 400 games against PUCT. Both algorithms use 32 evaluations for choosing their move. PUCT chooses the most simulated move and SHUSS chooses the only remaining move in $S_R$. The results for standard SHUSS with the usual exploration term $sc$ are given in Table \ref{shuss}. We can observe that the standard SHUSS is weaker than PUCT.

We also tested the two exploration terms discovered on the SHUSS dataset. The exploration term $pr + 2 \times sc \times sc$ is confronted to PUCT in Table \ref{shussPrior}. For all the PUCT constants we tested, SHUSS with the discovered exploration term is better than PUCT. The best result for PUCT is that SHUSS wins 51.00\% of its games. So we can say that the discovered exploration term made SHUSS competitive with PUCT.

The results for the exploration term found using Katago moves are given in Table \ref{shussKatago}. We can observe that this exploration term is much worse than the exploration term discovered on SHUSS moves with 128 evaluations. The results are even worse than standard SHUSS. These results illustrate that designing the target moves to be learned is very important. Counter intuitively it is better to learn moves from the same algorithm running longer than from the Katago moves that are better moves (Katago plays much better than SHUSS with 128 evaluations). This has some similarity to curriculum learning \cite{bengio2009curriculum}.

If we analyze the discovered exploration term $pr + 2 \times sc \times sc$, we see that when there are only a few playouts it takes into account the prior so as not to eliminate moves that have a great prior. When the number of playouts is greater, the $2 \times sc \times sc$ value becomes much greater than the prior and the moves are sorted according to the sum of their scores and not much according to their prior anymore. Table \ref{shuss64} gives the results of the exploration term with 64 playouts and the 6 best moves according to the prior. We can observe that the exploration is also competitive in this setting.

\section{Conclusion}

We presented a simple yet efficient method to find new exploration terms for Monte Carlo Tree Search. It uses sampling of mathematical expressions and a fast evaluation of the generated expressions. The generated exploration terms are simple. For search with a small number of evaluations, the method discovered an exploration term that works better than the usual exploration term for Sequential Halving. The discovered exploration term also beats the canonical PUCT algorithm for small equivalent search times. We also proposed the AMAF prior for sampling mathematical expressions. It reaches a score approximately 8 times faster than uniform sampling.

Our method to discover new exploration terms is simple, fast, general and empirically adapts the generated mathematical expressions to the problem at hand.

Future work involve accelerating the discovery of the expressions and applying the algorithm to other problems. It would also be interesting investigating the generation of more general expressions by evaluating them on more varied data, for example with different numbers of playouts or even for different games or problems.

From a more general point of view, Artificial Intelligence is becoming powerful enough to help discover new Artificial Intelligence algorithms. There are many further developments along the line of using Artificial Intelligence to improve Artificial Intelligence. 


\bibliographystyle{splncs04}
\bibliography{main}

\begin{thebibliography}{10}
\providecommand{\url}[1]{\texttt{#1}}
\providecommand{\urlprefix}{URL }
\providecommand{\doi}[1]{https://doi.org/#1}

\bibitem{bengio2009curriculum}
Bengio, Y., Louradour, J., Collobert, R., Weston, J.: Curriculum learning. In:
  Proceedings of the 26th annual international conference on machine learning.
  pp. 41--48 (2009)

\bibitem{BrownePWLCRTPSC2012}
Browne, C., Powley, E., Whitehouse, D., Lucas, S., Cowling, P., Rohlfshagen,
  P., Tavener, S., Perez, D., Samothrakis, S., Colton, S.: A survey of {M}onte
  {C}arlo tree search methods. {IEEE} Transactions on Computational
  Intelligence and {AI} in Games  \textbf{4}(1),  1--43 (Mar 2012).
  \doi{10.1109/TCIAIG.2012.2186810}

\bibitem{browne2020practical}
Browne, C., Stephenson, M., Piette, {\'E}., Soemers, D.J.: A practical
  introduction to the ludii general game system. In: Advances in Computer
  Games: 16th International Conference, ACG 2019, Macao, China, August 11--13,
  2019, Revised Selected Papers 16. pp. 167--179. Springer (2020)

\bibitem{Bruegmann1993MC}
Br\"{u}gmann, B.: {Monte Carlo Go}. Tech. rep., Max-Planke-Inst. Phys., Munich
  (1993)

\bibitem{cazenave2007evolving}
Cazenave, T.: Evolving monte carlo tree search algorithms. Dept. Inf., Univ.
  Paris  \textbf{8} (2007)

\bibitem{cazenave2010nested}
Cazenave, T.: Nested monte-carlo expression discovery. In: ECAI 2010, pp.
  1057--1058. IOS Press (2010)

\bibitem{cazenave2013monte}
Cazenave, T.: Monte-carlo expression discovery. International Journal on
  Artificial Intelligence Tools  \textbf{22}(01),  1250035 (2013)

\bibitem{cazenave2015generalized}
Cazenave, T.: Generalized rapid action value estimation. In: Yang, Q.,
  Wooldridge, M.J. (eds.) Proceedings of the Twenty-Fourth International Joint
  Conference on Artificial Intelligence, {IJCAI} 2015, Buenos Aires, Argentina,
  July 25-31, 2015. pp. 754--760. {AAAI} Press (2015)

\bibitem{cazenave2015sequential}
Cazenave, T.: Sequential halving applied to trees. {IEEE} Trans. Comput.
  Intell. {AI} Games  \textbf{7}(1),  102--105 (2015)

\bibitem{cazenave2015forecasting}
Cazenave, T., Hamida, S.B.: Forecasting financial volatility using nested monte
  carlo expression discovery. In: 2015 IEEE Symposium Series on Computational
  Intelligence. pp. 726--733. IEEE (2015)

\bibitem{cazenave2021cosine}
Cazenave, T., Sentuc, J., Videau, M.: Cosine annealing, mixnet and swish
  activation for computer {Go}. In: Advances in Computer Games. Springer (2021)

\bibitem{chen2024symbolic}
Chen, X., Liang, C., Huang, D., Real, E., Wang, K., Pham, H., Dong, X., Luong,
  T., Hsieh, C.J., Lu, Y., et~al.: Symbolic discovery of optimization
  algorithms. Advances in Neural Information Processing Systems  \textbf{36}
  (2024)

\bibitem{Coulom2006}
Coulom, R.: Efficient selectivity and backup operators in {Monte-Carlo} tree
  search. In: van~den Herik, H.J., Ciancarini, P., Donkers, H.H.L.M. (eds.)
  Computers and Games, 5th International Conference, {CG} 2006, Turin, Italy,
  May 29-31, 2006. Revised Papers. Lecture Notes in Computer Science,
  vol.~4630, pp. 72--83. Springer (2006)

\bibitem{danihelka2021policy}
Danihelka, I., Guez, A., Schrittwieser, J., Silver, D.: Policy improvement by
  planning with gumbel. In: International Conference on Learning
  Representations (2021)

\bibitem{fabiano2021sequential}
Fabiano, N., Cazenave, T.: Sequential halving using scores. In: Advances in
  Computer Games, pp. 41--52. Springer (2021)

\bibitem{fawzi2022discovering}
Fawzi, A., Balog, M., Huang, A., Hubert, T., Romera-Paredes, B., Barekatain,
  M., Novikov, A., R~Ruiz, F.J., Schrittwieser, J., Swirszcz, G., et~al.:
  Discovering faster matrix multiplication algorithms with reinforcement
  learning. Nature  \textbf{610}(7930),  47--53 (2022)

\bibitem{Finnsson2008}
Finnsson, H., Bj{\"o}rnsson, Y.: Simulation-based approach to general game
  playing. In: AAAI. pp. 259--264 (2008)

\bibitem{Finnsson2010}
Finnsson, H., Bj{\"{o}}rnsson, Y.: Learning simulation control in general
  game-playing agents. In: {AAAI} (2010),
  \url{http://www.aaai.org/ocs/index.php/AAAI/AAAI10/paper/view/1892}

\bibitem{Gelly2011AI}
Gelly, S., Silver, D.: {Monte-Carlo} tree search and rapid action value
  estimation in computer {Go}. Artif. Intell.  \textbf{175}(11),  1856--1875
  (2011)

\bibitem{hummel2022monte}
Hummel, A., Cazenave, T.: Monte carlo qubit routing. In: Quantum Machine
  Learning at ECML PKDD (2022)

\bibitem{illetskova2019nested}
Illetskova, M., Elnabarawy, I., Da~Silva, L.E.B., Tauritz, D.R., Wunsch, D.C.:
  Nested monte carlo search expression discovery for the automated design of
  fuzzy {ART} category choice functions. In: Proceedings of the Genetic and
  Evolutionary Computation Conference Companion. pp. 171--172 (2019)

\bibitem{SequentialHalving2013}
Karnin, Z.S., Koren, T., Somekh, O.: Almost optimal exploration in multi-armed
  bandits. In: ICML. pp. 1238--1246 (2013)

\bibitem{Kocsis2006}
Kocsis, L., Szepesv\'ari, C.: Bandit based {M}onte-{C}arlo planning. In: 17th
  European Conference on Machine Learning (ECML'06). LNCS, vol.~4212, pp.
  282--293. Springer (2006)

\bibitem{li2024discovering}
Li, Y., Li, W., Yu, L., Wu, M., Liu, J., Li, W., Hao, M., Wei, S., Deng, Y.:
  Discovering mathematical formulas from data via {GPT}-guided monte carlo tree
  search. arXiv preprint arXiv:2401.14424  (2024)

\bibitem{lu2021incorporating}
Lu, Q., Tao, F., Zhou, S., Wang, Z.: Incorporating actor-critic in monte carlo
  tree search for symbolic regression. Neural Computing and Applications
  \textbf{33},  8495--8511 (2021)

\bibitem{maes2013monte}
Maes, F., St-Pierre, D.L., Ernst, D.: Monte carlo search algorithm discovery
  for single-player games. IEEE Transactions on Computational Intelligence and
  AI in Games  \textbf{5}(3),  201--213 (2013)

\bibitem{mankowitz2023faster}
Mankowitz, D.J., Michi, A., Zhernov, A., Gelmi, M., Selvi, M., Paduraru, C.,
  Leurent, E., Iqbal, S., Lespiau, J.B., Ahern, A., et~al.: Faster sorting
  algorithms discovered using deep reinforcement learning. Nature
  \textbf{618}(7964),  257--263 (2023)

\bibitem{mehat2008monte}
M{\'{e}}hat, J., Cazenave, T.: {Monte-Carlo Tree Search} for general game
  playing. Univ. Paris  \textbf{8} (2008)

\bibitem{moudvrik2017algorithm}
Moud{\v{r}}{\'\i}k, J., K{\v{r}}en, T., Neruda, R.: Algorithm discovery with
  monte-carlo search: Controlling the size. In: 2017 IEEE 29th International
  Conference on Tools with Artificial Intelligence (ICTAI). pp. 390--395. IEEE
  (2017)

\bibitem{pepels2015sequential}
Pepels, T., Cazenave, T., Winands, M.H.M.: Sequential halving for partially
  observable games. In: Cazenave, T., Winands, M.H.M., Edelkamp, S., Schiffel,
  S., Thielscher, M., Togelius, J. (eds.) Computer Games - Fourth Workshop on
  Computer Games, {CGW} 2015, and the Fourth Workshop on General Intelligence
  in Game-Playing Agents, {GIGA} 2015, Held in Conjunction with the 24th
  International Conference on Artificial Intelligence, {IJCAI} 2015, Buenos
  Aires, Argentina, July 26-27, 2015, Revised Selected Papers. Communications
  in Computer and Information Science, vol.~614, pp. 16--29. Springer (2015)

\bibitem{pepels2014minimizing}
Pepels, T., Cazenave, T., Winands, M.H.M., Lanctot, M.: Minimizing simple and
  cumulative regret in monte-carlo tree search. In: Cazenave, T., Winands,
  M.H.M., Bj{\"{o}}rnsson, Y. (eds.) Computer Games - Third Workshop on
  Computer Games, {CGW} 2014, Held in Conjunction with the 21st European
  Conference on Artificial Intelligence, {ECAI} 2014, Prague, Czech Republic,
  August 18, 2014, Revised Selected Papers. Communications in Computer and
  Information Science, vol.~504, pp. 1--15. Springer (2014).
  \doi{10.1007/978-3-319-14923-3\_1},
  \url{https://doi.org/10.1007/978-3-319-14923-3\_1}

\bibitem{Pitrat68}
Pitrat, J.: Realization of a general game-playing program. In: {IFIP} Congress
  {(2)}. pp. 1570--1574 (1968)

\bibitem{pitrat2008step}
Pitrat, J.: A step toward an artificial artificial intelligence scientist
  (2008), {LIP6} Research Report

\bibitem{rosenhahn2023monte}
Rosenhahn, B., Osborne, T.J.: Monte carlo graph search for quantum circuit
  optimization. Physical Review A  \textbf{108}(6),  062615 (2023)

\bibitem{ruiz2024quantum}
Ruiz, F.J., Laakkonen, T., Bausch, J., Balog, M., Barekatain, M., Heras, F.J.,
  Novikov, A., Fitzpatrick, N., Romera-Paredes, B., van~de Wetering, J.,
  et~al.: Quantum circuit optimization with alphatensor. arXiv preprint
  arXiv:2402.14396  (2024)

\bibitem{sagri2024vision}
Sagri, A., Arjonilla, J., Saffidine, A., Cazenave, T.: Vision transformers for
  computer {Go}. In: EvoApps. Springer (2024)

\bibitem{schrittwieser2020mastering}
Schrittwieser, J., Antonoglou, I., Hubert, T., Simonyan, K., Sifre, L.,
  Schmitt, S., Guez, A., Lockhart, E., Hassabis, D., Graepel, T., et~al.:
  Mastering atari, go, chess and shogi by planning with a learned model. Nature
   \textbf{588}(7839),  604--609 (2020)

\bibitem{Silver2016MasteringTG}
Silver, D., Huang, A., Maddison, C.J., Guez, A., Sifre, L., van~den Driessche,
  G., Schrittwieser, J., Antonoglou, I., Panneershelvam, V., Lanctot, M.,
  Dieleman, S., Grewe, D., Nham, J., Kalchbrenner, N., Sutskever, I.,
  Lillicrap, T., Leach, M., Kavukcuoglu, K., Graepel, T., Hassabis, D.:
  Mastering the game of go with deep neural networks and tree search. Nature
  \textbf{529},  484--489 (2016)

\bibitem{silver2017mastering}
Silver, D., Hubert, T., Schrittwieser, J., Antonoglou, I., Lai, M., Guez, A.,
  Lanctot, M., Sifre, L., Kumaran, D., Graepel, T., Lillicrap, T.P., Simonyan,
  K., Hassabis, D.: Mastering chess and shogi by self-play with a general
  reinforcement learning algorithm. CoRR  \textbf{abs/1712.01815} (2017),
  \url{http://arxiv.org/abs/1712.01815}

\bibitem{silver2018general}
Silver, D., Hubert, T., Schrittwieser, J., Antonoglou, I., Lai, M., Guez, A.,
  Lanctot, M., Sifre, L., Kumaran, D., Graepel, T., et~al.: A general
  reinforcement learning algorithm that masters chess, shogi, and go through
  self-play. Science  \textbf{362}(6419),  1140--1144 (2018)

\bibitem{sun2022symbolic}
Sun, F., Liu, Y., Wang, J.X., Sun, H.: Symbolic physics learner: Discovering
  governing equations via monte carlo tree search. arXiv preprint
  arXiv:2205.13134  (2022)

\bibitem{swiechowski2023monte}
{\'S}wiechowski, M., Godlewski, K., Sawicki, B., Ma{\'n}dziuk, J.: Monte carlo
  tree search: A review of recent modifications and applications. Artificial
  Intelligence Review  \textbf{56}(3),  2497--2562 (2023)

\bibitem{wang2023automated}
Wang, P., Usman, M., Parampalli, U., Hollenberg, L.C., Myers, C.R.: Automated
  quantum circuit design with nested monte carlo tree search. IEEE Transactions
  on Quantum Engineering  (2023)

\bibitem{wu2019accelerating}
Wu, D.J.: Accelerating self-play learning in go. arXiv preprint
  arXiv:1902.10565  (2019)

\end{thebibliography}

\end{document}